\documentclass[11pt,letterpaper]{article}
\usepackage[margin=1in]{geometry}
\usepackage[T1]{fontenc}
\usepackage[hyphens]{url}
\usepackage{graphicx}
\usepackage[numbers,sort&compress]{natbib}
\usepackage{amsmath}
\usepackage{amssymb}
\usepackage{booktabs}
\usepackage{algorithm}
\usepackage{algorithmic}
\usepackage{authblk}

\date{}

\title{SkillShapley: Boundary-Adaptive Shapley Valuation\\for Skill Step Attribution in LLM Agents}

\author[1]{Chang Liu}
\author[1]{Yuqi Zhang}
\author[1]{Yiman Zhong}
\author[1]{Boyi Liu}
\author[1]{Hengjun Wang}
\author[2]{Shuyue Wei\thanks{Corresponding author: \texttt{weishuyue@sdu.edu.cn}}}

\affil[1]{Beihang University, Beijing, China}
\affil[2]{Shandong University, Jinan, China}

\begin{document}

\maketitle

\begin{abstract}
Agent skills are crucial external instructions that enable language agents to execute long procedural tasks such as coding or document processing.
Existing agent skills are primarily created through human manual crafting or agent execution traces, with limited understanding of how each step contributes to overall skill performance on specific tasks; i.e., \textit{there remains an open problem in quantifying the contribution of individual steps within an agent skill}.
To address this issue, we first model skill-step attribution as a Shapley value-based contribution estimation problem, and then propose \textsc{SkillShapley}, a step-level attribution framework for agent skills.
Notably, SkillShapley operates in two phases, motivated by key empirical insights, i.e., discretized benchmark rewards that create sharp performance cliffs, and step interactions that are largely additive rather than synergistic.
Specifically, it first identifies informative coalitional regions, and then adaptively samples new coalitions that can yield reusable marginal evidence.
Experiments on skills from the widely adopted SkillsBench demonstrate that our SkillShapley can effectively and efficiently identify high- or low-value skill steps, providing several key takeaways for agent skill creation.
\end{abstract}


\section{Introduction}

Large language models (LLMs) are increasingly deployed in the role of autonomous agents that execute multi-step procedures such as debugging, data analysis, and repository-level code changes~\citep{brown2020gpt3,achiam2023gpt4}.
A key mechanism for controlling such agents is the \emph{skill}: a step-by-step natural-language specification that guides the agent through a compound task.
Skills, tools, and workflow modules are widely used in systems such as SWE-agent~\citep{yang2024sweagent}, ReAct~\citep{yao2023react}, DSPy~\citep{khattab2024dspy}, and Toolformer~\citep{schick2023toolformer}, enabling substantial capability gains without further training the base model~\citep{mialon2023augmented}.

Despite their practical importance, skill design is still often done by trial and error.
Benchmarking can show \emph{how well} an end-to-end skill performs, but it does not explain \emph{which steps} are responsible for the performance or where redundancy lies.
Without fine-grained step-level attribution, practitioners often either over-specify skills, wasting context and cost, or prune blindly, risking sharp failures.

Existing methods have yet to offer a principled resolution to this problem.
Prompt optimization methods improve prompts or programs as a whole, prompt compression methods identify compressible tokens, and workflow optimizers improve execution structure, but none directly assigns value to individual steps within a fixed skill.
As a result, there is still no clear framework for asking which specific steps of an agent skill are truly doing the work.

We propose to treat a skill as a set of step ``players'' in a cooperative game and quantify each step's contribution with the Shapley value~\citep{shapley1953value}.
Shapley values provide a standard way to measure marginal contribution across contexts and have been successfully used for feature importance and data value~\citep{lundberg2017shap,ghorbani2019data}.
The central question is whether this attribution tool remains meaningful when the ``players'' are natural-language procedural steps.

Our motivation comes from an exact-reference exploration of low-step-count skills.
We compare Shapley values against two simple baselines used here: Individual scores, which evaluate each step in isolation, and Leave-One-Out (LOO) scores, which remove one step from the full skill.
In this setting, Shapley rankings produced differentiated step values, whereas Individual and LOO scores produced many ties.
Removing Shapley top-ranked steps also caused clear performance degradation.
This observation suggests that Shapley values can serve as a behaviorally meaningful target for skill-step attribution.
It also exposes the practical bottleneck: exact enumeration quickly becomes infeasible as the number of skill steps grows.

This work investigates three questions:
(1) whether Shapley values are behaviorally meaningful for natural-language skill steps;
(2) whether BAES can approximate exact Shapley rankings under a much smaller unique-configuration budget;
and (3) what the resulting attribution patterns reveal about skill pruning, revision, and creation.
We answer these questions with \textsc{SkillShapley} and a budgeted active approximation method called \textit{Boundary-Adaptive Edge Shapley} (BAES), which is designed for the skill regime where new configuration evaluations are expensive and cached one-flip comparisons can be reused.

\noindent\textbf{Contributions.}
We make three contributions.
(1) We formulate skill-step valuation as a cooperative-game attribution problem, where skill steps are players, retained-step subsets are coalitions, and benchmark performance is the utility.
(2) We introduce BAES, a cache-aware approximation method that combines warmup coverage with adaptive acquisition over reusable one-flip marginal edges.
(3) We show that BAES approximates Shapley values effectively in the skill setting, achieving better approximation with fewer unique configuration samples than competing estimators.

\section{Related Work}

\noindent\textbf{Agent Skills.}
LLM agents increasingly use external procedural knowledge, tools, and workflow modules to improve long-horizon task execution, including reasoning-and-acting frameworks~\citep{yao2023react}, tool-use models~\citep{schick2023toolformer,mialon2023augmented}, agent-computer interfaces~\citep{yang2024sweagent}, and declarative LM pipelines~\citep{khattab2024dspy}.
Recent SkillsBench formalizes agent skills as structured procedural knowledge and provides a benchmark of curated skills and deterministic verifiers across diverse tasks~\citep{li2026skillsbench}.
Skill optimization and refinement methods improve skill sets, refine LLM-authored skills, compress skill content, or compile skills across frameworks~\citep{nottingham2024sso,gautam2026skillaxe,gao2026skillreducer,ouyang2026skcc}.
However, existing skill-centered work generally treats skills as whole units, rather than assigning contribution scores to individual instruction steps inside a fixed skill.

\noindent\textbf{Instruction Attribution in LLMs.}
Beyond optimizing instructions as wholes, instruction attribution asks which parts of an LLM input or instruction affect model behavior.
Prior work probes prompt content~\citep{jiang2020prompt}, compresses prompts by identifying removable tokens~\citep{jiang2023llmlingua,jiang2024longllmlingua}, or assigns importance to tokens and spans~\citep{mosca2022shap,horovicz2024tokenshap}.
These methods provide fine-grained evidence about text units, but they do not directly address semantic procedure steps as the units of attribution.

\noindent\textbf{Shapley Values for Attribution.}
Shapley values provide a principled way to aggregate marginal contributions across contexts~\citep{shapley1953value}.
They are widely used for feature importance (e.g., SHAP~\citep{lundberg2017shap}) and data value (e.g., Data Shapley~\citep{ghorbani2019data}).
Because exact computation is exponential, many approximation methods have been proposed, including permutation sampling~\citep{castro2009polynomial}, stratified and uncertainty-aware sampling~\citep{castro2017improving,burgess2021approximating}, amortized estimators such as FastSHAP~\citep{jethani2022fastshap}, and broad empirical comparisons of Shapley estimators~\citep{chen2023algorithms}.
Most existing estimators target settings where coalition evaluation is relatively cheap compared with executing new LLM-agent skill variants.

Across these literatures, prior work optimizes skills or skill sets, but does not attribute value to individual steps inside a fixed skill under a benchmark distribution.
SkillShapley fills this gap by treating steps as Shapley players and by introducing BAES, a cache-aware approximation method tailored to the cost model and saturation patterns of skill evaluation.

\begin{figure}[t]
\centering
\includegraphics[width=0.98\textwidth]{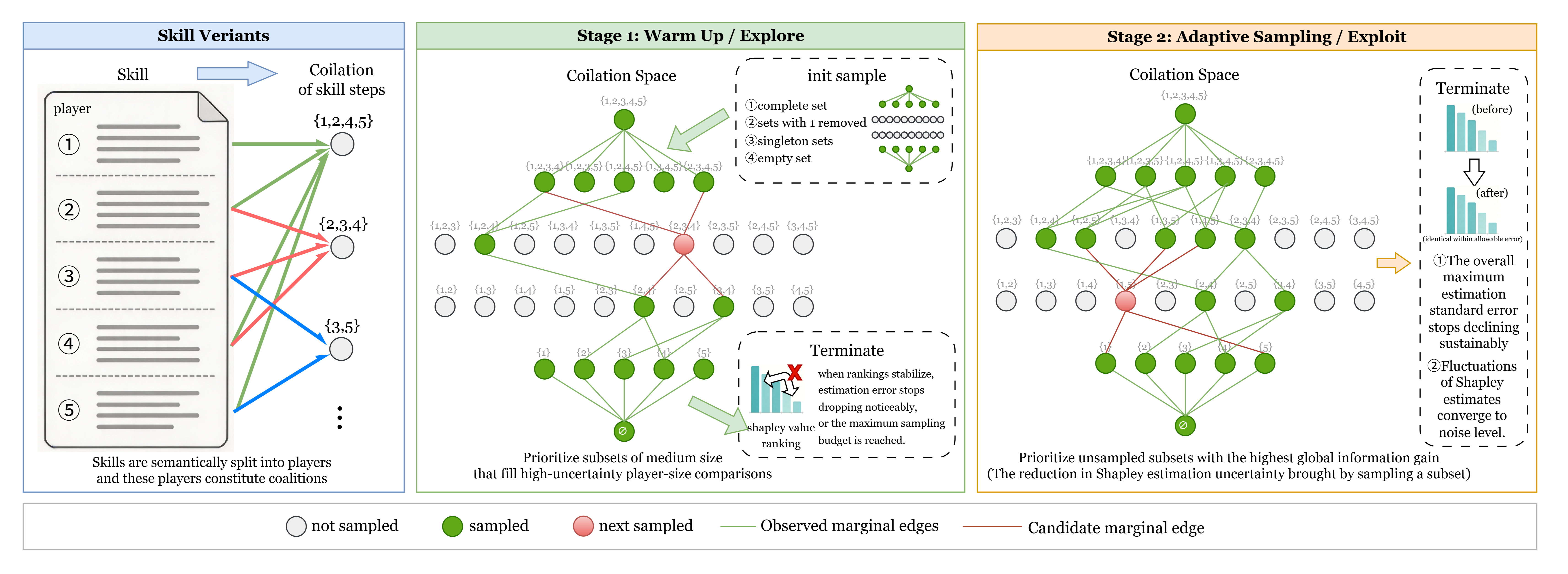}
\caption{Overview of BAES for budgeted skill-step attribution. The left panel shows how semantic instruction blocks define skill variants and enter the evaluation cache. The middle panel illustrates the warmup stage, which samples broadly across coalition sizes. The right panel illustrates the adaptive stage, which exploits the accumulated cache by selecting the next variant near high-priority regions with uncertain effects, under-sampled coverage, and many cached neighbors.}
\label{fig:method_overview}
\end{figure}

\section{Problem Formulation}
\label{sec:preliminaries}

We formulate skill-step attribution as a cooperative game because a skill is naturally composed of reusable instruction steps whose effects depend on which other steps are present.
Let a fixed skill be segmented into $n$ semantically coherent steps indexed by $N=\{1,\ldots,n\}$:
\begin{equation}
X=(x_1,\ldots,x_n),
\label{eq:skill_steps}
\end{equation}
where each index $i\in N$ is treated as one player.
The segmentation is done over the body of \texttt{skill.md}: blocks may correspond to task-entry guidance, decision rules, API examples, validation instructions, or common pitfalls, and tightly coupled snippets are not split when doing so would make a block semantically incomplete.
A subset of players $S\subseteq N$ corresponds to a skill variant that keeps exactly those steps and preserves their original order:
\begin{equation}
X_S=(x_i)_{i\in S,\ \mathrm{ordered\ as\ in}\ X}.
\label{eq:coalition_skill_variant}
\end{equation}
This subset construction gives the standard cooperative-game notion of a coalition: a coalition is not a different task or a rewritten prompt, but the same skill context with only a chosen set of steps retained.

To make the attribution well-defined, all factors other than the selected body steps are fixed across coalitions.
The frontmatter is always retained so the skill can still be discovered and loaded.
We also keep the non-target context, benchmark instances, and scoring rule fixed.
Let $\mathcal{B}=\{(q_j,y_j)\}_{j=1}^{M}$ be the fixed benchmark subset.
When the agent runs with skill variant $X_S$ on input $q_j$, it produces output $o_j(S)$.
The value of coalition $S$ is its measured benchmark utility:
\begin{equation}
v(S)
= \frac{1}{M}\sum_{j=1}^{M} r\!\left(o_j(S),y_j\right),
\qquad
v:2^N\rightarrow \mathbb{R},
\label{eq:coalition_utility}
\end{equation}
where $r(\cdot,\cdot)$ is the task scoring function.
In many skill benchmarks, $r$ is a binary success indicator, so $v(S)$ is an empirical success rate.

Under this cooperative-game view, skill-step attribution asks how much each player contributes to the utility function $v$ across possible coalitional contexts.
The desired output is a contribution vector
\begin{equation}
\boldsymbol{\alpha}=(\alpha_1,\ldots,\alpha_n)\in\mathbb{R}^n,
\label{eq:attribution_problem}
\end{equation}
where $\alpha_i$ summarizes the contribution of step $i$ to benchmark performance.
The practical constraint is that observing $v(S)$ requires executing an LLM agent with skill variant $X_S$ on benchmark instances, so each new coalition evaluation can be costly.

\section{Method}
\label{sec:method}

\subsection{Overview}

We instantiate the attribution vector in Eq.~\eqref{eq:attribution_problem} with Shapley values.
For step player $i\in N$, the target value is
\begin{equation}
\phi_i
= \sum_{S\subseteq N\setminus\{i\}}
\frac{|S|!(n-|S|-1)!}{n!}
\left[v(S\cup\{i\})-v(S)\right].
\label{eq:shapley}
\end{equation}
An equivalent size-stratified form is
\begin{equation}
\phi_i
= \frac{1}{n}\sum_{k=0}^{n-1}
\mathbb{E}\left[\Delta_i(S)\mid |S|=k,\ i\notin S\right],
\label{eq:size_stratified_shapley}
\end{equation}
where $\Delta_i(S)=v(S\cup\{i\})-v(S)$ is a one-step marginal contribution.
Eq.~\eqref{eq:size_stratified_shapley} partitions the target into player-size strata $(i,k)$ and separates where uncertainty lives across coalition sizes.

BAES is a budgeted active approximation method for Shapley values in the skill setting, where evaluation cost is dominated by new configurations, cached one-flip comparisons can be reused, and rewards are discrete and quickly flatten out.
The attribution target remains the standard Shapley value in Eq.~\eqref{eq:shapley}; BAES changes the sampling policy to decide which coalitions to evaluate when only a small number of configuration evaluations is affordable.
When used below, $g$ denotes the smallest observable reward increment.

BAES proceeds in two stages.
Algorithm~\ref{alg:baes} summarizes the complete procedure.
The warmup stage builds a cache that gives broad stratum coverage and identifies which strata remain uncertain.
The adaptive stage then repeatedly evaluates the single most informative unevaluated configuration, scored by the number of high-priority one-flip edges it forms with the cached configurations.
This separation between coverage and exploitation is important.
If BAES were purely greedy from the beginning, early random fluctuations could lock the sampling process onto a narrow set of strata.
The warmup stage prevents this failure mode by first creating enough local structure for acquisition scores to become meaningful.

\subsection{Observations}

BAES is derived from three observations about agent skill evaluation.
\textbf{First, configuration cost dominates edge cost:} the algorithm must decide which new configuration to evaluate so that it produces as many useful cached one-flip comparisons as possible.
\textbf{Second, rewards are discrete and noisy under limited repeated evaluation:} when each configuration is scored on only a small benchmark subset, the observed performance can change sharply from one coalition to another, making some strata much more uncertain than others.
\textbf{Third, rewards often flatten out over large parts of the coalition space:} when many neighboring coalitions have similar rewards, additional evaluations in those regions provide little new information.
Useful samples are therefore those that expose larger variation in marginal effects across player-size strata, suggesting adaptive sample allocation rather than uniform sampling.

Together, these observations motivate a two-stage strategy: a warmup stage that builds
coverage and identifies which strata remain uncertain, followed by an adaptive stage that
uses the remaining evaluations only where they continue to reduce uncertainty.
In effect, BAES treats cached configurations and their one-flip neighbors as a reusable
measurement structure rather than as a set of separate prompt variants.
The algorithm benefits when a new configuration simultaneously illuminates several
neighboring comparisons, especially around strata whose current estimates remain unstable.

\subsection{Two-Stage Procedure}

BAES maintains a cache $\mathcal{D}$ of evaluated configurations and records each observed one-flip comparison in its corresponding stratum $(i,k)$.
As the cache grows, BAES updates the empirical mean $\widehat{\mu}_{i,k}$, variance $\widehat{\sigma}^2_{i,k}$, and count $m_{i,k}$ of observed marginal effects within each stratum.
These statistics are used to decide where the next evaluation should go.
Specifically, BAES assigns each stratum an allocation score
\begin{equation}
a_{i,k}
= \frac{\sqrt{\widehat{\sigma}^2_{i,k}+\epsilon}}
{\sqrt{m_{i,k}+1}},
\label{eq:allocation_score}
\end{equation}
so high-variance and under-sampled strata receive more attention.
Here uncertainty is empirical rather than model-based: BAES does not predict the reward of an unevaluated coalition, but allocates new cache-aware evaluations toward strata whose observed marginal effects are still variable or poorly covered.
BAES also applies a coarse cached-reward weight
$b(y)=1+\max\{0,\operatorname{round}((1-y)/g)\}$, which gives higher priority to
configurations adjacent to low-reward cached states.
The warmup operator $\mathsf{TopPriority}$ ranks strata by
$a_{i,k}/\sqrt{m_{i,k}+1}$ and selects at most three unevaluated coalitions per
round that expose the highest-ranked strata while maximizing $A(C)$.
All ties are broken lexicographically by coalition bit string.

Warmup begins by evaluating anchor configurations that maximize potential edge reuse: the empty set, the full set, all singletons, and all $(n\!-\!1)$-subsets.
It then expands the cache with intermediate-size configurations selected by a greedy cache-aware rule that prefers candidates with many cached one-flip neighbors. Ties are resolved by a fixed deterministic ordering.
Each such configuration yields multiple marginal edges immediately, improving stratum coverage rapidly with only a small number of evaluations.
This cache-aware design creates more effective marginal-edge observations from the same configuration budget.
Warmup does not aim to produce the final best estimate; its purpose is to stabilize the relative priority ordering of strata.
Once the cache contains enough coverage, the algorithm can distinguish strata that are truly uncertain from strata that only appeared uncertain because they had not yet been observed.
The predicate $\mathsf{RankUnstable}$ compares successive allocation-score rankings with Kendall's $\tau$ and remains true until the exponential moving average of per-configuration $\tau$ improvement falls below $1/e$ of its observed peak after at least 60 observed strata, or until the warmup cap is reached.

After the cache reaches broad coverage, BAES transitions to adaptive acquisition.
At this stage, BAES scores an unevaluated configuration $C$ by
\begin{equation}
A(C)
= \sum_{i:\ C\triangle\{i\}\in\mathcal{D}}
a_{i,k}\cdot b\!\left(v(C\triangle\{i\})\right),
\label{eq:acquisition_score}
\end{equation}
where the sum runs over one-flip cached neighbors and $k$ is the context size associated with that edge.
BAES then evaluates the unevaluated configuration with the largest $A(C)$, adds it to the cache, and updates all affected strata statistics.
This directly targets configurations that simultaneously reduce uncertainty across multiple high-priority strata, rather than spreading samples uniformly across the lattice.
Because the acquisition score depends on the evolving cache, BAES is inherently adaptive: the value of evaluating a coalition depends not only on the coalition itself but also on which neighbors have already been evaluated.
This makes the sampling trajectory data-dependent, which is precisely what allows BAES to concentrate evaluations in the most informative parts of the search space.

\begin{algorithm}[tb]
\caption{Boundary-Adaptive Edge Shapley (BAES)}
\label{alg:baes}
\textbf{Input}: $N$ (step players), $v(\cdot)$ (evaluator), $g$ (reward granularity)\\
\textbf{Param}: $R$ (warmup cap), $B$ (total budget)\\
\textbf{Output}: $\widehat{\boldsymbol{\phi}}\in\mathbb{R}^{|N|}$ (step values)
\begin{algorithmic}[1]
\STATE $n\leftarrow |N|$ (player count); $\mathcal{D}\leftarrow\emptyset$ (cache)
\STATE $\mathcal{E}_{i,k}\leftarrow\emptyset,\ \forall i,k$ (edge strata)
\STATE $\mathcal{S}_0\leftarrow\{\emptyset,N\}\cup\{\{i\},N\setminus\{i\}:i\in N\}$ (anchors)
\STATE $\forall S\in\mathcal{S}_0:\ v(S)\leftarrow\mathsf{Eval}(S),\ \mathcal{D}\leftarrow\mathcal{D}\cup\{S\}$
\STATE $\mathcal{E}\leftarrow\mathsf{Edges}(\mathcal{D})$
\STATE $\mathcal{U}_{\rm rank}\leftarrow\mathsf{RankUnstable}(\mathcal{E})$ (rank unstable)
\WHILE{$\mathcal{U}_{\rm rank}\land |\mathcal{D}|<R$}
\STATE $\{\widehat{\mu},\widehat{\sigma}^2,m,a\}_{i,k}\leftarrow\mathsf{Update}(\mathcal{E},g)$
\STATE $\mathcal{C}_{\rm batch}^{\star}\leftarrow\mathsf{TopPriority}(\mathcal{D},\mathcal{E},a)$
\STATE $\mathcal{D}\leftarrow\mathcal{D}\cup\mathcal{C}_{\rm batch}^{\star};\ \mathcal{E}\leftarrow\mathsf{Edges}(\mathcal{D})$
\STATE $\mathcal{U}_{\rm rank}\leftarrow\mathsf{RankUnstable}(\mathcal{E})$
\ENDWHILE
\STATE $\mathcal{U}_{\rm uncert}\leftarrow\mathsf{UncertDec}(\mathcal{E},g)$ (uncertainty decreases)
\WHILE{$\mathcal{U}_{\rm uncert}\land |\mathcal{D}|<B$}
\STATE $a_{i,k}\leftarrow\sqrt{\widehat{\sigma}_{i,k}^{2}+\epsilon}/\sqrt{m_{i,k}+1},\ \forall i,k$
\STATE $C^\star\leftarrow\arg\max_{C\notin\mathcal{D}} A(C)$
\STATE $v(C^\star)\leftarrow\mathsf{Eval}(C^\star);\ \mathcal{D}\leftarrow\mathcal{D}\cup\{C^\star\}$
\STATE $\mathcal{E}\leftarrow\mathsf{Edges}(\mathcal{D});\ \mathcal{U}_{\rm uncert}\leftarrow\mathsf{UncertDec}(\mathcal{E},g)$
\ENDWHILE
\STATE $\widehat{\mu}_{i,k}\leftarrow\mathsf{Fill}(\widehat{\mu}_{i,k}),\ \forall(i,k):m_{i,k}=0$
\STATE \textbf{return} $\widehat{\phi}_i\triangleq n^{-1}\sum_{k=0}^{n-1}\widehat{\mu}_{i,k},\ \forall i\in N$
\end{algorithmic}
\end{algorithm}

\subsection{Stopping and Estimation}

To monitor whether additional evaluations still reduce uncertainty, BAES tracks a normalized standard-error signal over the stratum estimates.
The normalization uses the reward granularity $g$, which prevents the stopping rule from reacting to numerical changes below the resolution of the benchmark reward.
In Algorithm~\ref{alg:baes}, $\mathsf{UncertDec}$ denotes this check: it remains true while the recent slope of the normalized standard-error trajectory is still negative over a data-derived decorrelation window, with a minimum coverage requirement of 85 observed strata.
BAES stops when this uncertainty signal shows diminishing returns, or when a preset maximum sampling count is reached.
After stopping, it returns the size-stratified estimator
\begin{equation}
\widehat{\phi}_i
= \frac{1}{n}\sum_{k=0}^{n-1}\widehat{\mu}_{i,k}.
\label{eq:baes_estimator}
\end{equation}
If a small number of strata remain empty under a very small sampling count,
$\mathsf{Fill}$ replaces each missing $\widehat{\mu}_{i,k}$ with the mean of
observed stratum means at the same size $k$, or 0 if that size has no observed
stratum, so that $\widehat{\phi}$ is well-defined.

The target of BAES remains the standard Shapley value, but BAES is best understood as a
budgeted active approximation rather than as a claim of finite-sample unbiasedness under
every possible cooperative game.
Because the adaptive stage deliberately samples more often in uncertain or high-variation
strata, the empirical stratum means in Eq.~\eqref{eq:baes_estimator} are optimized for
low-budget ranking recovery rather than for uniform random sampling within each stratum.
When the budget is large enough that every player-size stratum is directly observed and
no fallback fill is used, the returned value reduces to the empirical
size-stratified average in Eq.~\eqref{eq:size_stratified_shapley}; under finite adaptive
budgets it should be interpreted as a biased approximation for ranking recovery.
Under the finite budgets relevant to LLM skill evaluation, we therefore evaluate BAES by
comparing it against exact Shapley references under matched unique-configuration budgets.
This matches the intended use case: the practitioner typically needs a low-cost estimate
that is close enough to guide which steps to preserve, inspect, or prune, not to report a
scalar value with negligible numerical error.

This stopping logic is designed for settings where rewards are coarse.
When $v(S)$ is measured on only a few benchmark instances, insisting on near-zero numerical error is unrealistic; a more useful criterion is whether additional evaluations are still changing the practically relevant ranking or only refining values below the reward granularity.

\begin{figure}[!t] \centering \includegraphics[width=0.98\textwidth]{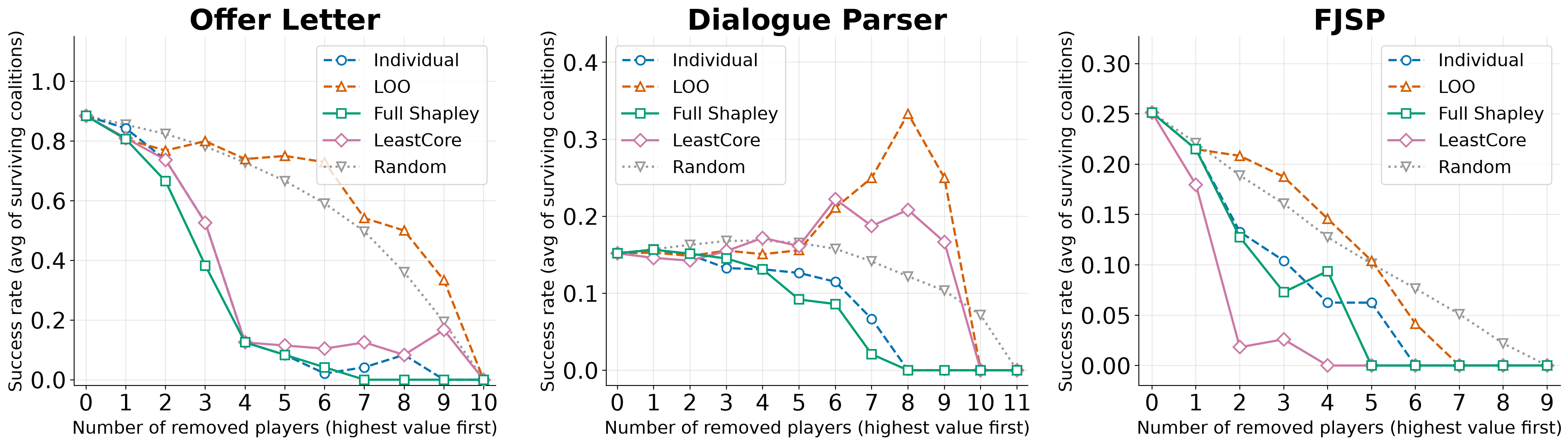} \caption{Experiment~1 removal validation on the three SkillsBench tasks. Each panel corresponds to one task. The x-axis is the number of removed players following each method's ranking. The y-axis is the mean success rate over all coalitions that do not contain the removed players. Curves compare simple baselines (Individual, Leave-One-Out, and Random Removal), LeastCore~\citep{maschler1979geometric}, and full Shapley~\citep{shapley1953value}. Stronger attribution should produce a sharper success-rate drop when its top-ranked players are removed.} \label{fig:rq1_removal_curves} \end{figure}
\section{Experiments}
\label{sec:results}

We evaluate SkillShapley along the three claims introduced above.
\textbf{First,} we test whether Shapley values are useful in the skill setting by comparing exact Shapley against simpler contribution measures and by validating the rankings through step-removal behavior.
\textbf{Second,} we test whether BAES is a better low-cost approximation to exact Shapley than other Shapley approximation methods under matched coalition-evaluation budgets.
\textbf{Third,} we use case studies from the resulting attribution profiles to extract guidance for skill pruning, revision, and creation.
This ordering is important: the exact-reference analysis first establishes that full Shapley values are behaviorally meaningful for skill steps; the approximation analysis then tests whether BAES can recover this validated target at lower cost; and the case study translates attribution patterns into editing guidance.

\subsection{Experimental Setup}
\label{sec:experiments}

All experiments use SkillsBench skills~\citep{li2026skillsbench}. We split each
\texttt{skill.md} into semantically coherent instruction blocks and treat each block as
one Shapley player. The frontmatter is always retained so that the skill remains
discoverable and loadable by the agent, and auxiliary resources such as reference files
and scripts are kept unchanged across all coalitions. For a coalition $S$, the
corresponding skill variant retains exactly the instruction blocks in $S$ while keeping
the task, system prompt, output format, benchmark instances, and scoring function fixed.
Thus changes in $v(S)$ are attributed to the retained instruction blocks rather than to
changes in evaluation data, package resources, or prompt paraphrasing.

Evaluating a coalition requires running the LLM skill on benchmark instances, which can
be slow and token-expensive. Therefore, the central cost in this setting is not the number
of arithmetic marginal comparisons, but the number of new skill variants that must be
executed by the model. We use the number of \emph{unique evaluated configurations} as the
primary cost measure and explicitly reuse cached coalition values whenever possible. All
approximation methods are compared under matched unique-configuration budgets. The agent
harness is OpenHands. All model calls use temperature $T=0$, and within each skill all
methods are evaluated on the same benchmark subset.

\subsection{Exact Shapley as Skill-Step Attribution}

The first experiment tests whether Shapley values are behaviorally meaningful for skill-step attribution.
It focuses on low-step-count SkillsBench skills for which exact enumeration is feasible: \textit{offer-letter-generator}, \textit{manufacturing-fjsp-optimization}, and \textit{dialogue parser}.
These skills need not share the same number of instruction blocks; each is evaluated over its own coalition lattice.
We compare exact Shapley with simple baselines, \textit{Individual}, \textit{Leave-One-Out}, and \textit{Random Removal}, and with \textit{LeastCore}~\citep{maschler1979geometric}, and validate the resulting rankings by measuring the utility degradation caused by removing top-ranked blocks.
We analyze these exact-reference skills to determine whether the cooperative-game view
produces a clearer and more actionable step ranking than simpler alternatives.

We evaluate attribution validity through top-ranked removal across all three
exact-reference tasks (Figure~\ref{fig:rq1_removal_curves}). If a method truly identifies
important steps, then deleting high-ranked steps should cause faster utility degradation
than deleting low-ranked or randomly selected steps. In Figure~\ref{fig:rq1_removal_curves},
the full Shapley removal curves drop fastest across the exact-reference tasks, supporting the claim that
Shapley values identify behaviorally important skill blocks more reliably than the simpler
baselines. We also report whether each method produces tied or diverse value profiles in
the accompanying interpretive text, but do not use a separate figure for this diagnostic.

This experiment establishes the target that BAES later approximates. Its role is therefore
foundational: before claiming that BAES is efficient, we verify that Shapley-based skill
rankings correspond to actual deletion risk in the SkillsBench setting.

\subsection{BAES Approximation Accuracy and Cost}

The second experiment tests whether BAES approximates the exact Shapley target more
effectively than other Shapley approximation methods. We use the exact Shapley values
from the low-step-count skills as references for budgeted approximation. All compared
methods estimate Shapley values: \textit{BAES}, \textit{Monte Carlo Shapley}~\citep{castro2009polynomial},
\textit{Quasi-Monte Carlo Shapley}~\citep{chen2023algorithms}, \textit{paired Monte Carlo Shapley}~\citep{chen2023algorithms}, and
\textit{size-$k$-truncated Shapley}, a simple baseline that keeps only marginal strata with coalition size at most $k$. The Monte Carlo variants follow the broader Shapley-estimation literature.
We compare them against the full Shapley reference under matched unique-configuration
budgets. This design isolates approximation quality from raw API cost: every method pays for
the same number of distinct skill variants, while BAES gains an advantage only if its
cache-aware acquisition selects more informative coalitions and reuses more marginal
edges from the same cache.

\begin{figure}[!t] \centering \includegraphics[width=0.98\textwidth]{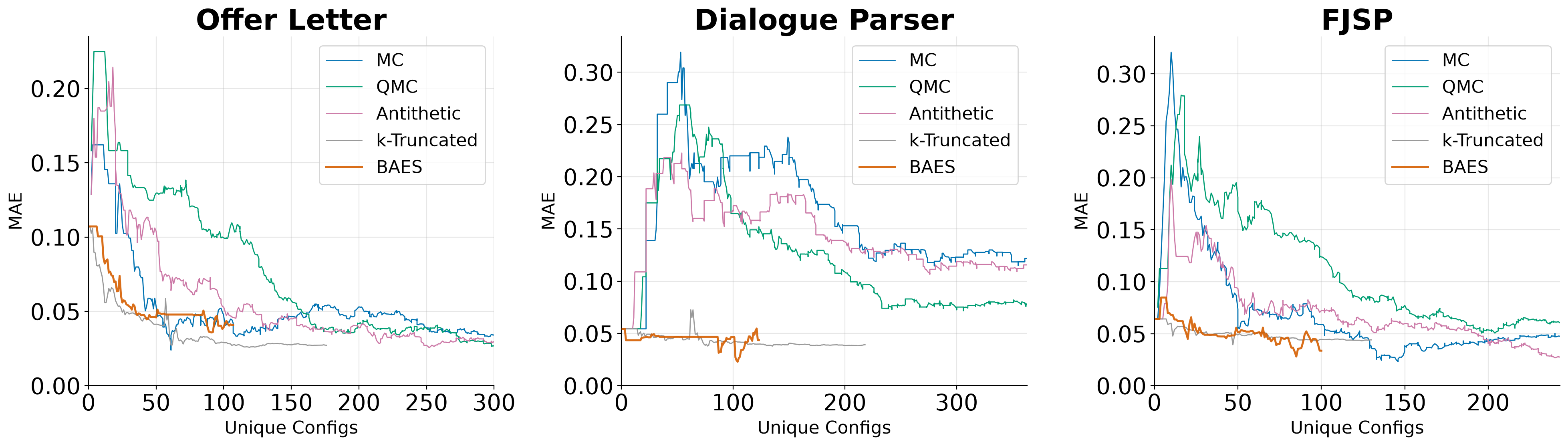} \caption{Experiment~2 approximation error under increasing unique-configuration budgets. The x-axis is the sampling budget, and the y-axis is error relative to full Shapley, aggregated over players. Lines compare BAES, Monte Carlo Shapley~\citep{castro2009polynomial}, Quasi-Monte Carlo Shapley~\citep{chen2023algorithms}, paired Monte Carlo Shapley~\citep{chen2023algorithms}, and size-$k$-truncated Shapley. Lower curves indicate faster convergence to the exact Shapley reference.} \label{fig:rq2_error_budget} \end{figure}

Figure~\ref{fig:rq2_error_budget} focuses on value error as a function of
unique-configuration budget. BAES reaches lower approximation error under smaller
budgets in the plotted comparison, indicating that cache-aware acquisition selects
more informative coalitions than the competing Shapley approximation baselines.

We also report a cache-efficiency diagnostic to separate BAES's sampling mechanism from raw
API cost. In our 10-player SkillsBench pilot, under the same 99 unique-configuration
budget, BAES Phase~1 yields 206 reusable one-flip marginal edges, whereas MC permutation
sampling yields 130 permutation marginal observations, only 115 of which are unique. This
supports the interpretation that BAES improves low-budget approximation partly by creating
more reusable marginal-edge evidence from the same number of evaluated skill variants.

\begin{figure}[!t]
\centering
\includegraphics[width=0.5\textwidth]{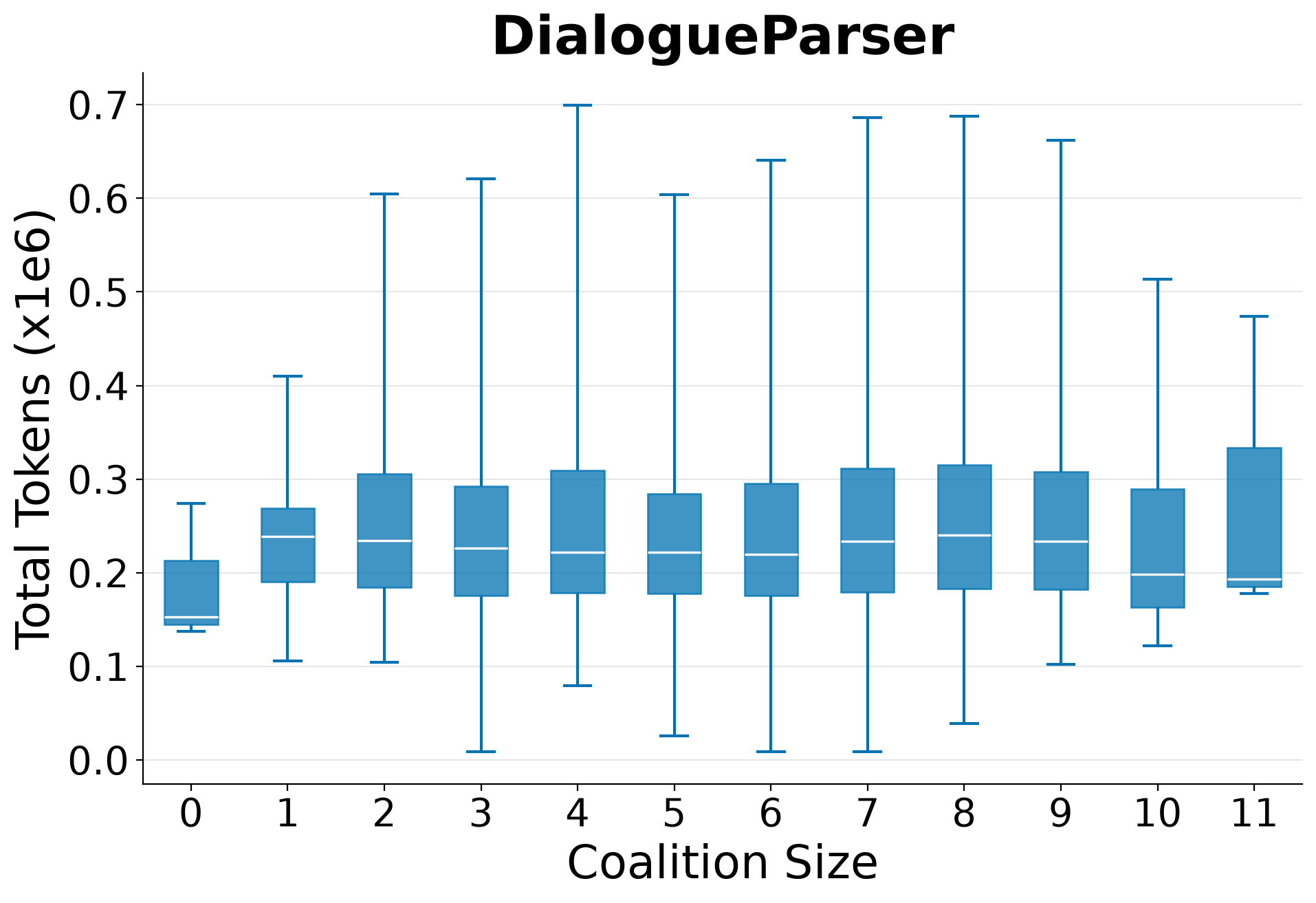}
\caption{Token-cost diagnostic for Dialogue Parser coalitions. The y-axis shows the total token count (input plus output tokens) for coalition variants with varying coalition sizes.}
\label{fig:token_cost_by_size}
\end{figure}

Figure~\ref{fig:token_cost_by_size} provides an auxiliary diagnostic on prompt cost for
the Dialogue Parser skill. Token count is not strongly determined by coalition size: many
coalition-size groups have similar median costs, and their ranges overlap substantially.
This suggests that token cost depends not only on how many instruction blocks are
retained, but also on which semantic blocks and auxiliary context are present. We
therefore treat attribution-guided editing as a way to remove low-value content, not as a
guarantee of proportional token savings.

\subsection{Case Study: Practical Guidance for Skill Creation and Modification}

Beyond validating attribution and approximation, SkillShapley provides multiple actionable takeaways for skill creation and modification.
Across the three cases, component value is better explained by procedural role than by surface form.
High-value steps tend to act as \emph{procedural bridges}: they connect a task condition to an executable decision rule, an API operation, or a constraint-aware fallback.
This pattern appears in different forms across the evaluated tasks.
In Offer Letter, high-value steps combine placeholder replacement with tables, nested tables, and headers or footers.
In Dialogue Parser, high-value steps connect the user scenario to concrete graph-building actions such as node construction, validation, or visualization.
In FJSP, high-value steps connect scheduling constraints to repair decisions, such as when to preserve the baseline machine and when to consider an alternative.
Low-value steps are often locally correct but action-incomplete: they provide background facts or isolated helper information without changing the agent's next decision.
Overall, these observations suggest that effective skills should be written as compact, decision-complete guidance units, rather than as shorter text or disconnected snippets.

\begin{center}
\setlength{\fboxsep}{5pt}
\fbox{\begin{minipage}{0.6\textwidth}
\scriptsize\raggedright
\textbf{High-value step: FJSP P9 (code block)} \hfill $\phi=0.1155$\\
\texttt{\# A naive ``always keep baseline machine'' can cause large start shifts.}\\
\texttt{This often reduces Shift\_L1 enough to pass tight budgets}\\
\texttt{without exploding machine changes. Use a simple trigger to}\\
\texttt{consider alternates only when it helps:}\\
\texttt{THRESH = 6 \# tune; small instances often 3--10 works}\\
\texttt{\# First try baseline machine}\\
\texttt{cand = best\_candidate\_restricted\_to([base\_m])}\\
\texttt{\# If shift is large and we still can change machines, search alternates}\\
\texttt{if (cand.start - base\_start) >= THRESH and mc\_used < max\_mc:}\\
\texttt{\ \ cand2 = best\_candidate\_over\_all\_allowed\_machines()}\\
\texttt{\ \ if cand2.start < cand.start:}\\
\texttt{\ \ \ \ cand = cand2}
\end{minipage}}

\vspace{0.6em}

\fbox{\begin{minipage}{0.6\textwidth}
\scriptsize\raggedright
\textbf{Low-value step: Offer Letter P2 (plain text)} \hfill $\phi=-0.0194$\\
This happens due to spell-check, formatting changes, or Word's internal XML structure.
\end{minipage}}
\end{center}

\noindent\textbf{Practical workflow.}
SkillShapley is most useful when users already have a fixed skill and want to know which steps are essential, redundant, or worth reinforcing.
Start from a candidate skill, use exact Shapley when feasible or BAES with a limited unique-configuration budget to estimate step values, propose one or two low-value deletions or reinforcements, and validate the edit using the same removal-curve protocol that underpins RQ1.
This turns skill editing into a measurable loop: hypothesize, edit, and re-evaluate, with attribution yielding a clear proposal mechanism rather than blind trial-and-error.

The same logic also supports skill creation, not only pruning. When attribution repeatedly
surfaces procedural bridges, those patterns indicate which conditions, decisions, or
fallbacks should be made explicit in future skills. Conversely, steps that are repeatedly
low-value across tasks and models become candidates for templating, compression, or
removal.
\section{Conclusion}

We introduced \textsc{SkillShapley}, a Shapley-value-based framework for step-level skill
attribution in LLM agents. Our SkillsBench experiments show that step importance is highly
heterogeneous within a skill, that the identity of critical steps depends on the task, and
that full Shapley produces clearer rankings and stronger top-ranked-removal effects than
the comparison methods.
Building on these observations, BAES leverages configuration-level caching and the
discrete, high-variance, and often saturated reward structure of skills to approximate
Shapley rankings efficiently with a small number of configuration evaluations.

Our exact-reference analysis focuses on low-step-count SkillsBench skills because exact
Shapley references are only practical in this regime. SkillShapley is also best suited to
contexts with a fixed player set and reasonably stable benchmark signals; dynamic-length
workflows and highly subjective assessment criteria make the underlying cooperative game
less well defined. Strongly coupled assembly-line workflows remain especially
challenging, because removing an intermediate step may collapse the entire pipeline and
make a large Shapley value reflect structural necessity rather than step usefulness.
Future work will expand exact anchors, improve uncertainty-aware editing decisions, and
extend step attribution to multi-skill agent pipelines and richer evaluation signals.

\bibliography{references}

\clearpage
\appendix

\section{Experiment Details}

For each evaluated skill, we manually segment \texttt{skill.md} into semantically
coherent instruction blocks, such as task-entry guidance, decision rules, API examples,
validation instructions, or common pitfalls. We do not split inside tightly coupled code
snippets or checklist items when doing so would make the block incomplete. Frontmatter is
retained in every coalition variant to preserve skill discovery and loading. Auxiliary
files, including references, scripts, templates, and test assets, are kept unchanged; only
selected instruction blocks in the body of \texttt{skill.md} are included or removed.
All coalition variants are generated by deleting steps without rewriting the
remaining steps, and the order of steps is kept across all coalitions.

The evaluated task--skill pairs are \textit{offer-letter-generator}/\textit{docx}
($n=10$, 1024 exact configurations),
\textit{manufacturing-fjsp-optimization}/\textit{fjsp-baseline-repair-with-downtime-and-policy}
($n=9$, 512 exact configurations), and
\textit{dialogue-parser}/\textit{dialogue-graph} ($n=11$, 2048 exact configurations).
Each configuration is evaluated on three benchmark instances, so the empirical reward
takes values in $\{0,1/3,2/3,1\}$ and $g=1/3$. The benchmark subset is fixed within each
skill and shared by all attribution and approximation methods. All coalitions are
executed with the OpenHands harness and temperature $T=0$. Approximation budgets follow
$B=3n^2$ and $R=\lfloor0.4B\rfloor$, giving $(B,R)=(300,120)$, $(243,97)$, and
$(363,145)$ for the three pairs. Metrics are player-wise MAE against the exact Shapley
vector and removal success under the top-ranked deletion protocol.

\section{BAES Implementation Details}

For each player-size stratum $(i,k)$, BAES records $m_{i,k}$ observed marginal effects and computes
\begin{equation}
\begin{aligned}
\widehat{\mu}_{i,k}
= \frac{1}{m_{i,k}}\sum_{j=1}^{m_{i,k}}\Delta_i(S_j),
\\
\widehat{\sigma}^2_{i,k}
= \frac{1}{m_{i,k}-1}\sum_{j=1}^{m_{i,k}}
\left(\Delta_i(S_j)-\widehat{\mu}_{i,k}\right)^2.
\end{aligned}
\label{eq:app_stratum_statistics}
\end{equation}
When $m_{i,k}\leq 1$, we set the empirical variance to the default unseen-stratum
variance used by the allocation score.
The cached-reward weight used in the acquisition score is
\begin{equation}
b(v_{\text{cached}})
= 1 + \max\!\left(0,\ \mathrm{round}\!\left(\frac{1-v_{\text{cached}}}{g}\right)\right),
\label{eq:app_cached_reward_weight}
\end{equation}
where $g$ is the reward granularity.
For stopping, BAES monitors the normalized standard error
\begin{equation}
\begin{aligned}
\mathrm{NSE}_{i,k}
= \frac{\mathrm{SE}(\widehat{\mu}_{i,k})}{\max\left(|\widehat{\phi}_{\max}|,\ g\right) + \delta},
\\
\text{where}\quad
\mathrm{SE}(\widehat{\mu}_{i,k}) = \frac{\widehat{\sigma}_{i,k}}{\sqrt{m_{i,k}}}.
\end{aligned}
\label{eq:app_nse}
\end{equation}
Here $\widehat{\phi}_{\max}=\max_{i\in N}|\widehat{\phi}_i|$ for the current
estimate and $\delta=10^{-8}$.

\noindent\textbf{Default operators.}
BAES evaluates $\emptyset$, $N$, all $\{i\}$, and all $N\setminus\{i\}$ before greedy
warmup. The allocation score is
$a_{i,k}=\sqrt{\widehat{\sigma}_{i,k}^{2}+(g/10)^2}/\sqrt{m_{i,k}+1}$; unseen strata
use the median observed positive variance, or $g^2/4$ if no positive variance exists.
$\mathsf{TopPriority}$ ranks strata by $a_{i,k}/\sqrt{m_{i,k}+1}$, inspects the top 30
strata, and adds up to three unevaluated coalitions that expose those strata and
maximize $A(C)$. The acquisition score $A(C)$ sums
$a_{i,k}b(v(C\triangle\{i\}))$ over cached one-flip neighbors, with lexicographic
coalition bit strings breaking ties. $\mathsf{RankUnstable}$ stops warmup when the
exponential moving average of $\Delta\tau/\Delta|\mathcal{D}|$ falls below $1/e$ of its
observed peak, with $\tau>0$ and at least 60 observed strata; otherwise it stops at $R$.
The EMA half-life is three warmup rounds. $\mathsf{UncertDec}$ stops adaptive sampling
when the recent NSE slope over a decorrelation window is non-decreasing and at least 85
strata are observed; otherwise it stops at $B$. The decorrelation lag is the first lag
whose autocorrelation is not significantly positive under the one-sided 95\% rule
$r<1.645/\sqrt{T-\ell}$. $\mathsf{Fill}$ uses the mean of observed
$\widehat{\mu}_{j,k}$ at the same size $k$ for empty $(i,k)$, and uses 0 if no stratum
at size $k$ has been observed. Recorded diagnostics are stratum coverage, reusable edge
counts, and the NSE trajectory.

\end{document}